\documentclass{llncs}

\usepackage[width=122mm,left=12mm,paperwidth=146mm,height=193mm,top=12mm,paperheight=217mm]{geometry}

\usepackage{epsfig}
\usepackage{graphicx}
\usepackage{amsmath}
\usepackage{amssymb}
\usepackage{scrextend}
\usepackage{multirow}
\usepackage[english]{babel}
\usepackage{subcaption}
\usepackage{paralist, tabularx}
\usepackage{enumitem}   
\usepackage{graphicx}
\usepackage{booktabs}
\usepackage{caption}
\usepackage{subcaption}
\usepackage{xspace}
\usepackage{tikz}
\usepackage{pgfplots}
\usetikzlibrary{spy,calc}

\makeatletter
\DeclareRobustCommand\onedot{\futurelet\@let@token\@onedot}
\def\@onedot{\ifx\@let@token.\else.\null\fi\xspace}

\def\eg{\emph{e.g}\onedot} 
\def\ie{\emph{i.e}\onedot}

\def\etal{\emph{et al}\onedot}
\makeatother

\usepackage[pagebackref,breaklinks,colorlinks]{hyperref}

\usepackage[capitalize]{cleveref}
\crefname{section}{Sec.}{Secs.}
\Crefname{section}{Section}{Sections}
\Crefname{table}{Table}{Tables}
\crefname{table}{Tab.}{Tabs.}

\newif\ifblackandwhitecycle
\gdef\patternnumber{0}

\pgfkeys{/tikz/.cd,
    zoombox paths/.style={
        draw=orange,
        very thick
    },
    black and white/.is choice,
    black and white/.default=static,
    black and white/static/.style={ 
        draw=white,   
        zoombox paths/.append style={
            draw=white,
            postaction={
                draw=black,
                loosely dashed
            }
        }
    },
    black and white/static/.code={
        \gdef\patternnumber{1}
    },
    black and white/cycle/.code={
        \blackandwhitecycletrue
        \gdef\patternnumber{1}
    },
    black and white pattern/.is choice,
    black and white pattern/0/.style={},
    black and white pattern/1/.style={    
            draw=white,
            postaction={
                draw=black,
                dash pattern=on 2pt off 2pt
            }
    },
    black and white pattern/2/.style={    
            draw=white,
            postaction={
                draw=black,
                dash pattern=on 4pt off 4pt
            }
    },
    black and white pattern/3/.style={    
            draw=white,
            postaction={
                draw=black,
                dash pattern=on 4pt off 4pt on 1pt off 4pt
            }
    },
    black and white pattern/4/.style={    
            draw=white,
            postaction={
                draw=black,
                dash pattern=on 4pt off 2pt on 2 pt off 2pt on 2 pt off 2pt
            }
    },
    zoomboxarray inner gap/.initial=5pt,
    zoomboxarray columns/.initial=2,
    zoomboxarray rows/.initial=2,
    subfigurename/.initial={},
    figurename/.initial={zoombox},
    zoomboxarray/.style={
        execute at begin picture={
            \begin{scope}[
                spy using outlines={%
                    zoombox paths,
                    width=\imagewidth / \pgfkeysvalueof{/tikz/zoomboxarray columns} - (\pgfkeysvalueof{/tikz/zoomboxarray columns} - 1) / \pgfkeysvalueof{/tikz/zoomboxarray columns} * \pgfkeysvalueof{/tikz/zoomboxarray inner gap} -\pgflinewidth,
                    height=\imageheight / \pgfkeysvalueof{/tikz/zoomboxarray rows} - (\pgfkeysvalueof{/tikz/zoomboxarray rows} - 1) / \pgfkeysvalueof{/tikz/zoomboxarray rows} * \pgfkeysvalueof{/tikz/zoomboxarray inner gap}-\pgflinewidth,
                    magnification=3,
                    every spy on node/.style={
                        zoombox paths
                    },
                    every spy in node/.style={
                        zoombox paths
                    }
                }
            ]
        },
        execute at end picture={
            \end{scope}
            \node at (image.north) [anchor=north,inner sep=0pt] {\subcaptionbox{\label{\pgfkeysvalueof{/tikz/figurename}-image}}{\phantomimage}};
            \node at (zoomboxes container.north) [anchor=north,inner sep=0pt] {\subcaptionbox{\label{\pgfkeysvalueof{/tikz/figurename}-zoom}}{\phantomimage}};
     \gdef\patternnumber{0}
        },
        spymargin/.initial=1ex,
        zoomboxes xshift/.initial=1,
        zoomboxes right/.code=\pgfkeys{/tikz/zoomboxes xshift=1},
        zoomboxes left/.code=\pgfkeys{/tikz/zoomboxes xshift=-1},
        zoomboxes yshift/.initial=0,
        zoomboxes above/.code={
            \pgfkeys{/tikz/zoomboxes yshift=1},
            \pgfkeys{/tikz/zoomboxes xshift=0}
        },
        zoomboxes below/.code={
            \pgfkeys{/tikz/zoomboxes yshift=-1},
            \pgfkeys{/tikz/zoomboxes xshift=0}
        },
        caption margin/.initial=1ex,
    },
    adjust caption spacing/.code={},
    image container/.style={
        inner sep=0pt,
        at=(image.north),
        anchor=north,
        adjust caption spacing
    },
    zoomboxes container/.style={
        inner sep=0pt,
        at=(image.north),
        anchor=north,
        name=zoomboxes container,
        xshift=\pgfkeysvalueof{/tikz/zoomboxes xshift}*(\imagewidth+\pgfkeysvalueof{/tikz/spymargin}),
        yshift=\pgfkeysvalueof{/tikz/zoomboxes yshift}*(\imageheight+\pgfkeysvalueof{/tikz/spymargin}+\pgfkeysvalueof{/tikz/caption margin}),
        adjust caption spacing
    },
    calculate dimensions/.code={
        \pgfpointdiff{\pgfpointanchor{image}{south west} }{\pgfpointanchor{image}{north east} }
        \pgfgetlastxy{\imagewidth}{\imageheight}
        \global\let\imagewidth=\imagewidth
        \global\let\imageheight=\imageheight
        \gdef\columncount{1}
        \gdef\rowcount{1}
        
    },
    image node/.style={
        inner sep=0pt,
        name=image,
        anchor=south west,
        append after command={
            [calculate dimensions]
            node [image container,subfigurename=\pgfkeysvalueof{/tikz/figurename}-image] {\phantomimage}
            node [zoomboxes container,subfigurename=\pgfkeysvalueof{/tikz/figurename}-zoom] {\phantomimage}
        }
    },
    color code/.style={
        zoombox paths/.append style={draw=#1}
    },
    connect zoomboxes/.style={
    spy connection path={\draw[draw=none,zoombox paths] (tikzspyonnode) -- (tikzspyinnode);}
    },
    help grid code/.code={
        \begin{scope}[
                x={(image.south east)},
                y={(image.north west)},
                font=\footnotesize,
                help lines,
                overlay
            ]
            \foreach \x in {0,1,...,9} { 
                \draw(\x/10,0) -- (\x/10,1);
                \node [anchor=north] at (\x/10,0) {0.\x};
            }
            \foreach \y in {0,1,...,9} {
                \draw(0,\y/10) -- (1,\y/10);                        \node [anchor=east] at (0,\y/10) {0.\y};
            }
        \end{scope}    
    },
    help grid/.style={
        append after command={
            [help grid code]
        }
    },
}

\newcommand\phantomimage{%
    \phantom{%
        \rule{\imagewidth}{\imageheight}%
    }%
}
\newcommand\zoombox[2][]{
    \begin{scope}[zoombox paths]
        \pgfmathsetmacro\xpos{
            (\columncount-1)*(\imagewidth / \pgfkeysvalueof{/tikz/zoomboxarray columns} + \pgfkeysvalueof{/tikz/zoomboxarray inner gap} / \pgfkeysvalueof{/tikz/zoomboxarray columns} ) + \pgflinewidth
        }
        \pgfmathsetmacro\ypos{
            (\rowcount-1)*( \imageheight / \pgfkeysvalueof{/tikz/zoomboxarray rows} + \pgfkeysvalueof{/tikz/zoomboxarray inner gap} / \pgfkeysvalueof{/tikz/zoomboxarray rows} ) + 0.5*\pgflinewidth
        }
        \edef\dospy{\noexpand\spy [
            #1,
            zoombox paths/.append style={
                black and white pattern=\patternnumber
            },
            every spy on node/.append style={#1},
            x=\imagewidth,
            y=\imageheight
        ] on (#2) in node [anchor=north west] at ($(zoomboxes container.north west)+(\xpos pt,-\ypos pt)$);}
        \dospy
        \pgfmathtruncatemacro\pgfmathresult{ifthenelse(\columncount==\pgfkeysvalueof{/tikz/zoomboxarray columns},\rowcount+1,\rowcount)}
        \global\let\rowcount=\pgfmathresult
        \pgfmathtruncatemacro\pgfmathresult{ifthenelse(\columncount==\pgfkeysvalueof{/tikz/zoomboxarray columns},1,\columncount+1)}
        \global\let\columncount=\pgfmathresult
        \ifblackandwhitecycle
            \pgfmathtruncatemacro{\newpatternnumber}{\patternnumber+1}
            \global\edef\patternnumber{\newpatternnumber}
        \fi
    \end{scope}
}

\begin{document}
\pagestyle{headings}
\mainmatter
\title{Unsupervised Video Interpolation by Learning Multilayered 2.5D Motion Fields} 

%
\titlerunning{FVI by 2.5D Motion Fields}
\author{Ziang Cheng\inst{1} \and
Shihao Jiang\inst{2} \and
Hongdong Li\inst{1}}
%
%
\institute{The Australian National University \and Cognitive Computing Lab, Baidu Research\\
\email{\{ziang.cheng,hongdong.li\}@anu.edu.au}}


\maketitle

\begin{abstract}
The problem of video frame interpolation is to increase the temporal resolution of a low frame-rate video, by interpolating novel frames between existing temporally sparse frames. This paper presents a self-supervised approach to video frame interpolation that requires only a single video. We pose the video as a set of layers. Each layer is parameterized by two implicit neural networks --- one for learning a static frame and the other for a time-varying motion field corresponding to video dynamics. Together they represent an occlusion-free subset of the scene with a pseudo-depth channel. To model inter-layer occlusions, all layers are lifted to the 2.5D space so that the frontal layer occludes distant layers. This is done by assigning each layer a depth channel, which we call `pseudo-depth', whose partial order defines the occlusion between layers. The pseudo-depths are converted to visibility values through a fully differentiable SoftMin function so that closer layers are more visible than layers in a distance. On the other hand, we parameterize the video motions by solving an ordinary differentiable equation (ODE) defined on a time-varying neural velocity field that guarantees valid motions. This implicit neural representation learns the video as a space-time continuum, allowing frame interpolation at any temporal resolution. We demonstrate the effectiveness of our method on real-world datasets, where our method achieves comparable performance to state-of-the-arts that require ground truth labels for training.
\end{abstract}

\section{Introduction}

The problem of video frame interpolation (VFI) is to synthesize novel in-between frames given a set of temporally sparse input frames. 
Most existing approaches address the frame interpolation problem
with optical flow, which models the motion between the given frames
and interpolate the new frame linearly with the estimated motion. While
high quality results have already been shown in existing works 
\cite{bao2019depth,huang2020rife}, they often require a large training
dataset with ground-truth to first train an optical flow network to make 
frame interpolation successful. However, such dataset may not always be available, particularly 
for data domains where rendering each frame is costly and VFI is highly desired (\eg CGI, cartoons).

In this paper, 
we propose to address the VFI problem from a different paradigm. 
Instead of estimating motions between two frames and predicting new
frames from the estimated motion, we propose to directly model a video
sequence with neural implicit representations. 
The only 
required knowledge is a single sequence of temporally sparse video frames 
and their timestamps, from which an implicit neural representation of the 
video is extracted. Such representation defines a video as a space-time 
continuum of brightness values, allowing frames to be interpolated at any 
resolution. Departing from existing methods based on either forward or 
backward warping, we learn the video motion as invertible neural flow 
fields that are temporally consistent by construction, and the motion 
displacements are obtained by solving a system of ordinary differential 
equations (ODE) with respect to time. Inspired by image layers featured in image 
editing software, we stack multiple video layers to account for motion 
occlusion, which otherwise has been a long standing problem for frame 
interpolation. This paradigm sets forth a natural representation that is 
simple to implement and easy to train.

\begin{figure}
    \centering
    \includegraphics[width=\textwidth]{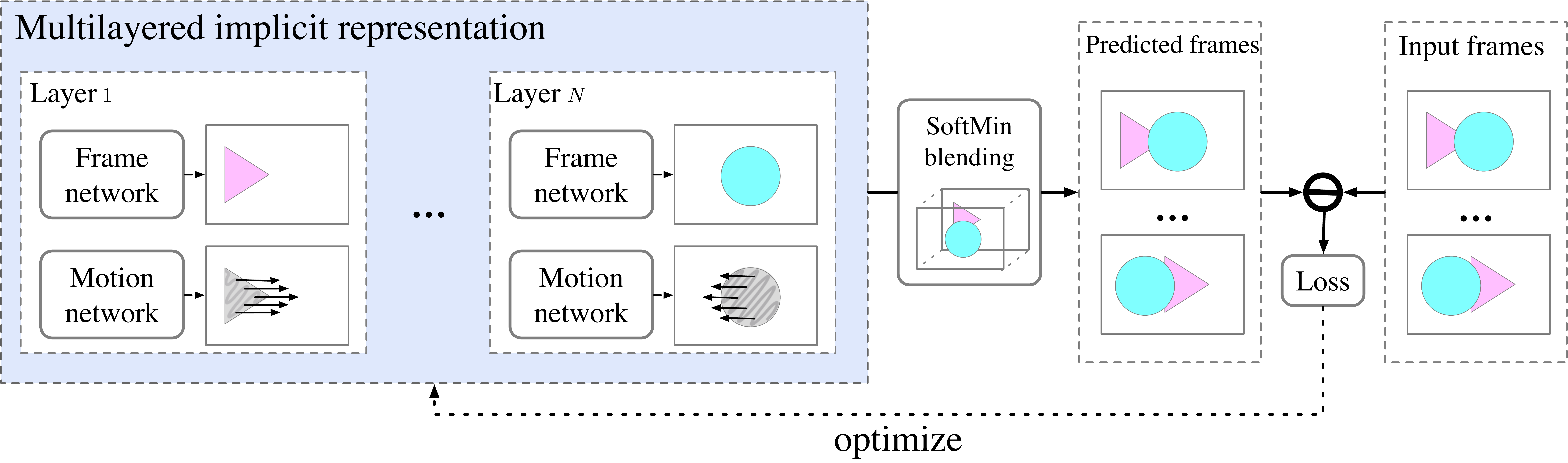}
    \caption{\small \textbf{Overview of our method.} We use a multilayered implicit representation where each video layer is modeled by two coordinate-based neural nets, one for a static frame and the other for the corresponding motion flow field. The layers are combined through a SoftMin aggregation function that blends all layers into video frames. The networks represent video as a space-time continuum and are supervised by input video and a few regularization terms.}
    \label{fig:pipeline}
\end{figure}

Figure~\ref{fig:pipeline} shows an overview of our solution for a toy video. The video is separated into multiple video layers. Each layer is characterized by a static frame and a corresponding dynamic flow field that moves and/or deforms the frame through time. The static frame and its motion field are modeled by two coordinate-based neural nets respectively. To account for video occlusions, all layers feature a depth channel. We call this channel pseudo-depth, since it encodes not necessarily the actual depth but is instead only used for rendering layers. We further differentiate occlusion boundaries by converting pseudo-depths to visibility values, so that closer depths correspond to higher opacity/visibility (see also Figure\ref{fig:layer_view}). By compositing multiple layers in this 2.5D space, our solution is able to establish complex video motion and occlusion solely supervised by input frames and a few regularization terms.

To summarize, our paper makes following contributions:
\begin{itemize}
    \item An implicit neural representation is learned from a single multi-frame video. There is no need for collecting a large scale dataset or ground truth labels.
    \item Video motion is encoded by a flow field that is smooth and always temporally consistent.
    \item By stacking multiple video layers, our method is occlusion-aware by construction. We do not explicitly supervise occlusion map/mask, instead the networks learn occlusion boundaries on their own.
\end{itemize}

\section{Related work}

\paragraph{Flow-free VFI:}
Meyer \etal \cite{meyer2015phase} first proposed using a phase-based approach to allow in-between images to be generated by simple per-pixel phase modification, without the need for explicit correspondence estimation. Long \etal \cite{long2016learning} proposed the first deep learning approach to address the unsupervised learning problem of image matching, by formulating the image matching problem also as a frame interpolation problem. Flow-free VFI approaches often adopt the kernel-based method \cite{niklaus2017ada,niklaus2017adasep,cheng2020video},  where a convolution kernel is used to synthesize the intermediate frame from two input frames.  Niklaus \etal \cite{niklaus2017ada} first proposed the adaptive convolution approach where a kernel map is first predicted by a CNN and is then used to convolve the two input images to synthesize the intermediate image. Niklaus \etal \cite{niklaus2017adasep} improves the computational efficiency of adaptive convolution by adopting 1D separable filters in addressing the frame interpolation problem. Cheng and Chen \cite{cheng2020video} improves the performance of prior works \cite{niklaus2017adasep,niklaus2017ada} by introducing  deformable separable convolution, which adaptively estimate kernels, offsets and masks to
allow the network to obtain information from more faraway pixels with smaller kernel size. Their approach handles large motions especially well. Choi \etal \cite{choi2020channel} recently proposed a flow-free approach where only the
PixelShuffle operation and a sequence of residual channel attention blocks are employed to synthesize the intermediate frame. 

\paragraph{Flow-based VFI:}
Apart from the flow-free approaches, a number of works do incorporate optical flow
estimation as a key module in addressing the frame interpolation problem. Liu \etal
\cite{liu2017video} pioneered in deep flow-based VFI by predicting voxel flow from
input images and use trilinear interpolation to generate the output frame. Niklaus \etal \cite{niklaus2018context} proposed adding in warped context features when synthesizing the final intermediate frame and observed performance improvement. Super-slomo  \cite{jiang2018super} first proposed an approach to perform variable-length multi-frame interpolation compared to single-frame interpolation. Followed by Super-slomo, Reda \etal \cite{reda2019unsupervised} proposed an approach to learn frame interpolation on low frame-rate videos in an unsupervised manner, by a cycle consistency loss. Bao \cite{bao2019memc} proposed to combine flow-based and kernel-based methods into one motion estimation motion compensation (MEMC) module. Bao \etal also proposed DAIN \cite{bao2019depth}, which address the occlusion problem in frame interpolation by exploring the depth information. When estimating project flow vectors, input flow vectors are inversely weighted by depth, signalling objects with smaller depth values occlude those with larger depth values. Different from linear motion assumption in existing works, Xu \etal \cite{Xu2019Quadratic} adopts a quadratic motion model by using four consecutive frames. Liu \etal \cite{liu2020enhanced} enhanced QVI \cite{Xu2019Quadratic} by employing rectified quadratic flow predictions.  She \etal \cite{shen2020blurry} jointly addressed the video deblur and frame interpolation problem with a pyramid module and a recurrent network. Park \etal \cite{park2020bmbc} proposed a new network architecture to predict bilateral motions and interpolate frames  with dynamic blending filters. Niklaus \etal \cite{niklaus2020softmax} proposed an differentiable forward warping method with softmax splatting and achieves SOTA results on the frame interpolation task. Recently, RIFE \cite{huang2020rife} proposed IFNet to directly predict intermediate flows, as opposed to first estimating bi-directional flows then inverting to obtain intermediate flows. RIFE achieves much faster inference speed compared to previous approaches as well as SOTA results. A knowledge distillation scheme is adopted to supervise IFNet. Sim \etal \cite{sim2021xvfi} propose a multi-scale network, dubbed XVFI-Net, to handle extremely large displacement in video motions. They combine a cascade scale-invariant module and a fixed scale module to approximate optical flow and obtained high quality results on their large motion dataset. Finally, Park \etal \cite{park2021asymmetric} estimate an asymmetric bilateral motion field to warp consecutive frames to an intermediate frame.

\paragraph{Layer-based video reasoning: }
The concept of video layer has long been used for object segmentation and extraction \cite{smith2004layered,criminisi2006bilayer}. Alayrac~\etal~\cite{alayrac2019visual} explored video layers for isolating transmitted frames from reflection and/or shadow on the transmission medium. Akimoto \etal \cite{akimoto2020fast} separate videos into semi-transparent regions for video re-coloring. Ke \etal \cite{ke2021deep} and Liu \etal \cite{liu2020learning} use a bi-layer representation to handle occlusions between foreground/background objects. 

\paragraph{Our Approach:}
Our approach towards the frame interpolation problem is principally different from
existing approaches. All existing works require a large high frame-rate dataset for training, and many need an optical flow module trained with ground truth annotations. On the contrary, our approach is trained without ground truth. The only other unsupervised method by Reda \etal \cite{reda2019unsupervised} adopts cycle-consistency as supervision signal and still require a large database for training. In contrast, our solution is cycle-consistent by construction, and does not need a training database. Instead, only a single video clip is required. Two MLPs are optimized on the given video clip which implicitly represent both the brightness and motion cues. 

Existing work use optical flow to represent motion between frames at distinct time instances, and often use linear approximation to represent motion between frames. In this work, we learn the motion of the video as flow fields, which have the capacity to represent instantaneous motion velocities at any time instance. Our approach also does not require forward or backward warping to synthesize the new frame, which is known to produce artifacts. 

With occlusion handling, our approach is depth-aware similar to DAIN \cite{bao2019depth}. However, DAIN uses a separate pre-trained depth network to help weight contributing optical flow vectors. In our case, we construct multiple layers representing different depth levels. Our core idea is to let neural networks learn a per-layer pseudo-depth distribution that explains occlusion boundaries, in a way similar to the multiplane image (MPI) representation in the field of novel-view synthesis \cite{zhou2018stereo,tucker2020single}. This approach differs from layer-based CNNs, in that we do not directly supervise layer separations, but rather allow implicit representations to supervise themselves solely from input video. Unlike previous methods that handle occlusion by a binary mask, our solution explicitly predicts both occluder and the occluded parts of scene, and is able to model semi-transparent objects, moving shadows and reflections.

\section{Neural video representation}

Our neural video representations express a video as a static frame moved by a dynamic motion field. The frame itself and its motion field are modeled by two implicit neural networks. More specifically, to enforce a valid motion field, we define it as a bijective flow field of motion trajectories. Such definition, however,  prohibits occlusion and disocclusion, since under either circumstance objects can disappear behind or emerge from the occluder where no trajectory can be traced. To overcome this limitation, we stack multiple layers of neural representations in the 2.5D space, so that the layer at the smallest depth occludes all other layers. 

We shall first examine an occlusion-free case in Section \ref{sec:single_layer}. Section \ref{sec:multi_layer} provides a detailed explanation of our multi-layer design, and we leave the motion field definition last in Section \ref{section:motion_fields}.


\subsection{Single video layer for the occlusion-free}\label{sec:single_layer}

Suppose the scene being filmed is occlusion-free, it is assumed that the video can be approximated by a single layer, with a static canonical frame and a time-varying motion field imposed on such frame. More formally, any occlusion-free video, defined as a mapping $V:\{u,v,t\}\mapsto \{R,G,B\}$, is a function composition of a canonical frame $F:\{u_F,v_F\}\mapsto \{R,G,B\}$ at some fixed timestamp $t_F$, and a location mapping (warping) $M:\{u_0,v_0,t_0,t_1\}\mapsto \{u_1,v_1\}$ that moves the pixel at $(u_0,v_0)$ to location $(u_1,v_1)$ during time $t_0$ to $t_1$,
\begin{equation}
I(u,v,t) = F\big( M(u,v,t,t_F) \big).
\end{equation}

We employ two neural networks to approximate the canonical frame and its motion field respectively. A feed forward MLP is trained for the canonical frame representation $F$. On the other hand, the mapping $M$ needs to be a valid motion field. We enforce this condition via a Recurrent weighted ResNet structure as discussed later. 

\subsection{Occlusion by stacking layers in the 2.5D}\label{sec:multi_layer}

\begin{figure}[!htb]
    \centering
    \includegraphics[width=0.9\textwidth]{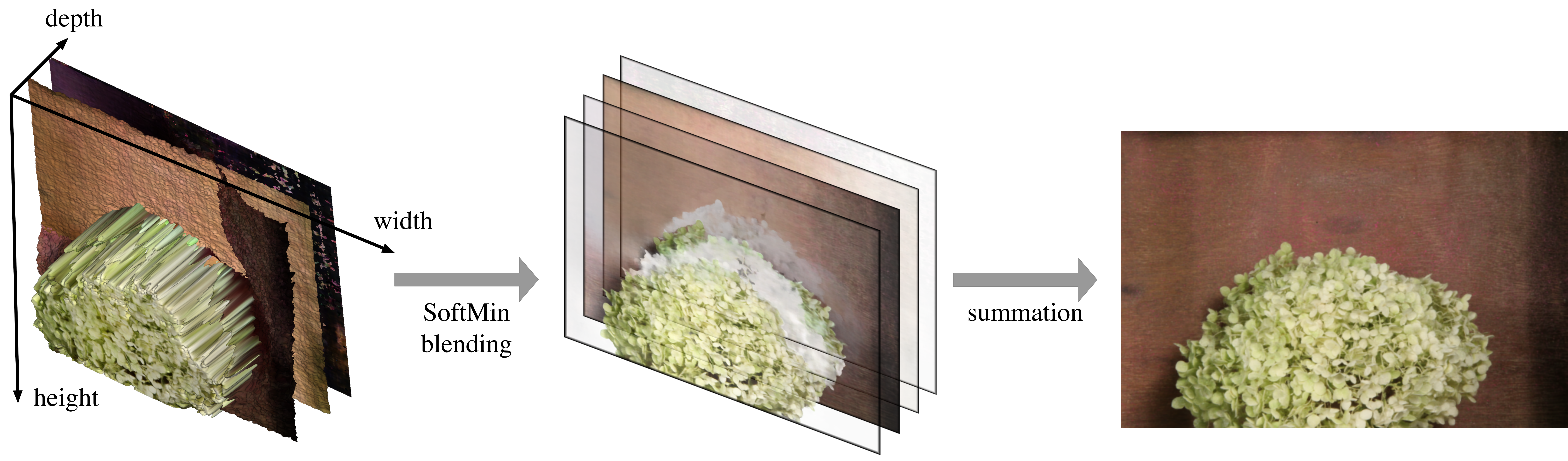}
    \caption{\textbf{Layer-based occlusion reasoning.} We lift video layers to 2.5D by learning a pseudo-depth channel for each layer. A fully differentiable SoftMin blending is applied to the depth dimension to convert depth to normalized visibility/opacity values. Closer depth leads to greater visibility, effectively blocking layers further away in occluding regions.}
    \label{fig:layer_view}
\end{figure}

Unlike previous methods that handle occlusion by binary masks, here we account for occlusion by stacking multiple video layers, each depicting an occlusion-free subset of scene whose motion can be characterized by a velocity field. Compared to the occlusion mask approach, this representation is more complete, since it models not only the occluding but also the occluded parts of the scene. And by making the video layers semi-transparent, we are able to represent non-opaque occluders, moving shadows and reflections. The layers span a 3-dimensional space with the front-most layer occluding the rest everywhere, and occlusion boundaries correspond to locations where layers intersect. Notably, this design makes the networks occlusion-aware without explicit occlusion handling (\eg predicting forward/backward occlusion mask as done in \cite{bao2019memc,sim2021xvfi}), and the networks learn to predict occlusion boundaries completely on their own.

Figure~\ref{fig:layer_view} illustrates our layer-based approach for occlusion handling. To account for motion occlusions in the 3-dimensional space, for each video layer $I_i$ we assign an additional pseudo-depth channel to its canonical frame $F_i\{u_0,v_0\}\mapsto \{R,G,B,D\}$. That is, each layer is now lifted to a 2.5D space, represented by an RGB-D frame with a motion field $V_i$ bound to it, and the order of depth defines inter-layer occlusions. The pseudo-depths themselves do not have to be consistent with actual depth of view, as long as the order of depth faithfully represent scene occlusions. More specifically, it suffices that the occluding layer has smaller pseudo-depth values than occluded layers' in occluding regions. To render the final RGB video from all layers, we use the `SoftMin' function to replace the hard, one-hot occlusion, converting their depth to visibility values within range $(0,1)$. RGB values from individual layers are blended according to the corresponding visibilities.
\begin{equation}
    I(\cdot) = \frac{\sum_i exp\big(-\gamma I_i^{D}(\cdot)\big) I_i^{RGB}(\cdot)}{\sum_i exp\big(-\gamma I_i^{D}(\cdot)\big) }.
\end{equation}
Unlike the hard occlusion, this soft blending is fully differentiable and can further explain semi-transparent occluders and dynamic shadowing/lighting effects. We add a parameter $\gamma=5$ to control the trade-off between differentiability and hardness of occlusion. Again, we note that the pseudo-depth of layers may not be consistent with the true depth of scene. Instead, their values are only used for rendering video frames. 

\begin{figure}
    \centering
    \resizebox{!}{0.3\textwidth}{\begin{tikzpicture}[
squarednode/.style={rectangle, rounded corners=3, draw=black, very thick, minimum height=1cm, align=center},
delta_t_wrap/.style={rectangle, rounded corners=3, draw=black, thick, align=center, dotted},
dashedsquarednode/.style={rectangle, dashed, rounded corners=6, draw=black, thick, minimum size=5mm},
arrow_text/.style={thick, font=\small},
text_box/.style={rectangle, font=\small},
background_box/.style={rounded corners=6, dashed, thick, color=green(pigment)},
]

\fill[fill=red!5, rounded corners=12](-3.25,-2.75) rectangle (0.5, 3);
\fill[fill=red!5](0,-2.75) rectangle (0.5, 3);
\fill[fill=blue!5](0.5, 3) rectangle (1, -2.75);
\fill[fill=blue!5, rounded corners=12](0.5, 3) rectangle (3.85, -2.75);
\node[dashedsquarednode,minimum height=4mm] (input) at (-4,0) {$u_F,v_F$};
\node[squarednode, minimum width=1.1cm, minimum height=3cm] (pos_enc)     at(-2.5, 0)   {\\\\\\\\ Pos. \\ Enc.};

\draw[very thick](-2.5,0.6) circle (0.4);
\draw[blue, very thick] (-2.85, 0.6) sin (-2.675,0.8) cos (-2.5,0.6) sin (-2.325,0.4) cos (-2.15,0.6);

\node[squarednode, minimum width=4.5cm, minimum height=0.7cm, rotate=90] (fc_layer1)     at(-1.25, 0)   {FC\hspace{0.3cm}+\hspace{0.3cm}Sinusoidal};


\node[squarednode, minimum width=4.5cm, minimum height=0.7cm, rotate=90] (fc_layer2)     at(-0.25, 0)   {(...)};

\node[squarednode, minimum width=4.5cm, minimum height=0.7cm, rotate=90] (fc_layer3)     at(1.25, 0)   {(...)};

\node[squarednode, minimum width=4.5cm, minimum height=0.7cm, rotate=90] (fc_layer4)     at(2.25, 0)   {FC\hspace{0.3cm}+\hspace{0.3cm}Sinusoidal};

\node[squarednode, minimum width=3.5cm, minimum height=0.7cm, rotate=90] (fc_layer5)     at(3.25, 0)   {FC\hspace{0.3cm}+\hspace{0.3cm}Sigmoid};

\node[dashedsquarednode,minimum height=4mm] (output) at (5,0) {$R,G,B,D$};

\node[rectangle] at(-1.25, 2) {$256$};
\node[rectangle] at(-0.25, 2) {$256$};
\node[rectangle] at(1.25, 2) {$256$};
\node[rectangle] at(2.25, 2) {$256$};
\node[rectangle] at(3.25, 1.5) {$4$};

\draw[->, arrow_text, very thick] (input.east) -- (pos_enc.west);
\draw[->, arrow_text, very thick] (pos_enc.east) -- (fc_layer1.north);
\draw[->, arrow_text, very thick] (fc_layer1.south) -- (fc_layer2.north);
\draw[->, arrow_text, very thick] (fc_layer2.south) -- (fc_layer3.north);
\draw[->, arrow_text, very thick] (fc_layer3.south) -- (fc_layer4.north);
\draw[->, arrow_text, very thick] (fc_layer4.south) -- (fc_layer5.north);
\draw[->, arrow_text, very thick] (fc_layer5.south) -- (output.west);
\draw[decoration={brace,raise=5pt},decorate, thick] (-1.5,2.2) -- node[above=6pt] {$4$ layers} (0,2.2);
\draw[decoration={brace,raise=5pt},decorate, thick] (1,2.2) -- node[above=6pt] {$4$ layers} (2.5,2.2);
\end{tikzpicture}}
    \caption{\textbf{The static frame network $F_i$.} The network features position encoding and SIREN modules for learning higher frequency variations \cite{tancik2020fourier,sitzmann2020implicit}. The frontal subnet, colored in red, is shared by all video layers.}
    \label{fig:fnet}
    \vspace*{-2em}
\end{figure}
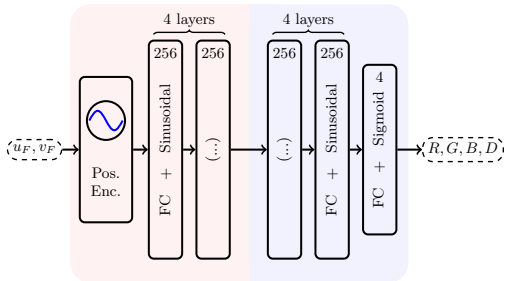

We use an MLP with sinusoidal activation function \cite{sitzmann2020implicit} for frame representation network $F_i$, as illustrated in Figure~\ref{fig:fnet}. We found that a Fourier position encoding layers further helps accelerate training \cite{tancik2020fourier}. To obtain a more meaningful layer-wise separation, all MLPs $F_i$ share a front subnet (colored in red in Figure~\ref{fig:fnet}) so that they have some common knowledge about the video scene to work with. The last layer of the MLP is activated by sigmoid function to constrain the range of RGBD values.

The minimum number of video layers required to account for all motion occlusions is determined by the length of topological order of occluders. For a detailed discussion the reader is referred to supplementary materials.  Empirically, however, a few layers often suffice to recreate occlusions in most real world videos, as validated later in Section~\ref{section:resuslts}. In this paper we use a constant 4 layers representation unless otherwise stated.

\subsection{Video dynamics as time-varying motion fields}\label{section:motion_fields}

Importantly, the location mapping must describe a valid motion field parameterized by time. That is to say, mapping $[u_1,v_1]^\top=M_i(u_0,v_0,t_0,t_1)$ defines a motion trajectory from $(u_0,v_0)$ to $(u_1,v_1)$ through time $t_0$ to $t_1$, that is temporally consistent:
\begin{enumerate}
    \item If $t_0=t_1$ then $u_0=u_1$ and $v_0=v_1$.
    \item Trajectories are forward consistent: for any $t\in[t_0,t_1]$, $M\big(M_i(u_0,v_0,t_0,t),t,t_1\big)=[u_1,v_1]^\top$.
    \item Trajectories are backward consistent with respect to time, \ie $M(u_1,v_1,t_1,t_0)=[u_0,v_0]^\top$.
\end{enumerate}
Intuitively, these conditions state that scene components in the video must travel smoothly (as opposed to instantly jumping around), and the location mapping vary continuously in space and is backward consistent in time. Without these restrictions, the mapping $M_i$ would not generally describe a valid motion field. One such example would be a mapping that takes pixel coordinates $(u,v,t)$ to locations on a color palette frame $F$ -- while such mapping could still reproduce the entire video, it does not represent underlying scene motions and cannot be used for interpolating frames.

To enforce a valid motion field, instead of directly regressing the location mapping $M$ itself, we learn an ordinary differentiable equation (ODE) of a time-varying velocity field that corresponds to video motions. Such velocity field $V:\{u,v,t\}\mapsto \mathrm{R}^2$ governs the time derivatives of location mapping $M$, yielding the following ODE,
\begin{equation}
    V\big(M(u_0,v_0,t_0,t),t\big) = \frac{\partial M(u_0,v_0,t_0,t)}{\partial t} .
\end{equation}
Conversely, given a velocity field, we may compute its location mapping by integration over time.
\begin{equation}
    M(u_0,v_0,t_0,t_1) = \int_{t_0}^{t_1} V\big(M(u_0,v_0,t_0,t),t\big) dt + [u_0,v_0]^\top \label{integration}
\end{equation}
As long as $V$ is smooth, the resulting location mapping $M$ is always a valid motion field. In fact, one can further show that such integration $M(\cdot,\cdot,t_0,t_1)$ yields a group of diffeomorphisms parameterized by time variables $t_0$ and $t_1$. We refer the reader to supplementary materials for a proof. A detailed theoretical treatment on diffeomorphisms and velocity fields can also be found in \cite{lee2013smooth}.

\begin{figure}[!htb]
\centering
\resizebox{!}{0.3\textwidth}{\begin{tikzpicture}[
squarednode/.style={rectangle, rounded corners=3, draw=black, very thick, minimum height=1cm, align=center},
delta_t_wrap/.style={rectangle, rounded corners=3, draw=black, thick, align=center, dotted},
dashedsquarednode/.style={rectangle, dashed, rounded corners=6, draw=black, thick, minimum size=5mm},
arrow_text/.style={thick, font=\small},
text_box/.style={rectangle, font=\small},
background_box/.style={rounded corners=6, dashed, thick, color=green(pigment)},
]

\fill[fill=blue!5, rounded corners=12](-3.25,-2.75) rectangle (4, 3);

\node[dashedsquarednode, minimum height=0.45cm] (input)     at(-4, 0.25)   {$u_i,v_i$};
\node[dashedsquarednode, minimum height=0.45cm] (t_input)     at(-3.5, -0.25)   {$t_i$};

\node[squarednode, minimum width=4.5cm, minimum height=0.7cm, rotate=90] (fc_layer1)     at(-2, 0)   {FC\hspace{0.3cm}+\hspace{0.3cm}L.ReLU};

\node[squarednode, minimum width=4.5cm, minimum height=0.7cm, rotate=90] (fc_layer2)     at(-1, 0)   {FC\hspace{0.3cm}+\hspace{0.3cm}L.ReLU};

\node[squarednode, minimum width=3cm, minimum height=0.7cm, rotate=90] (fc_layer3)     at(0, 0)   {FC\hspace{0.3cm}+\hspace{0.3cm}tanh};

\node[rectangle, font=\huge, inner sep=0pt] (concat_syl) at(1.45, 0) {$\oplus$};

\node[dashedsquarednode, minimum height=0.45cm] (output)     at(3, 0)   {$u_{i+1},v_{i+1}$};
\node[rectangle] at(-2, 2) {$256$};
\node[rectangle] at(-1, 2) {$256$};
\node[rectangle] at(0, 1.25) {$2$};

\draw[->, arrow_text,very thick] (t_input.east) -- ($(fc_layer1.north)+(0,-0.25)$);
\draw[->, arrow_text,very thick] ($(input.east)+(-0.1,0)$) -- ($(fc_layer1.north)+(0,0.25)$);
\draw[->, arrow_text,very thick] (fc_layer1.south) -- (fc_layer2.north);
\draw[->, arrow_text,very thick] (fc_layer2.south) -- (fc_layer3.north);
\draw[->, arrow_text,very thick] (fc_layer3.south) -- ($(concat_syl.west)+(0.05,0)$);
\path[arrow_text] (fc_layer3.south) -- ($(concat_syl.west)+(0.05,0)$) node[midway,above] {$\times\Delta t$};

\draw[->, arrow_text,very thick] ($(concat_syl.east)+(-0.07,0)$) -- (output.west);

\draw[->, arrow_text,very thick] (input.south) to ($(input)+(0,-2.75)$) to ($(concat_syl)+(0,-2.5)$) to ($(concat_syl.south) + (0, 0.02)$);

\draw[->, arrow_text, color=blue,very thick] (output.north) to ($(output)+(0,2.5)$) to ($(input)+(0,2.25)$) to (input.north);
\path[arrow_text] (5.3, 2.35) -- ($(input)+(0,2.35)$) node[midway,above] {$t_{i+1}=t_i+\Delta t,\ i=i+1$};

\end{tikzpicture}}
\caption{\textbf{Structure of motion field network $M_i$.} We use a recurrent weighted residual architecture to enforce validity of motion fields. The velocity outputs are weighted by time step $\Delta t$ then added back to the current location, constituting one discrete step along the motion trajectory. The recurrent link is colored in blue.}\label{fig:mnet}
\end{figure}
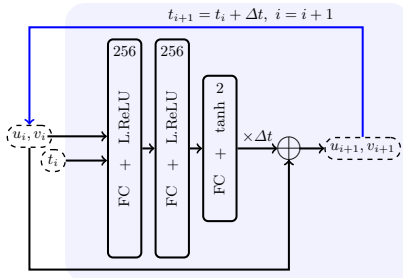

In this paper the velocity field is modeled by a light-weighted MLP. For computing the integration in Eq.\ref{integration}, we use a discrete approximation known as the Euler method, by taking small time steps of size $\Delta t$ along the trajectories.
\begin{align}
    M(\cdot,\cdot,t_0,t_{i+1}) &= M(\cdot,\cdot,t_0,t_{i}) + \Delta t\ V\big(M(\cdot,\cdot,t_0,t_{i}), t_{i}\big),\nonumber \\ &\text{where}\; t_{i+1} = t_{i}+\Delta t,\; i=0,1,2,... \label{eq:discrete}
\end{align}
Using above formula, one may compute the mapping $M$ iteratively by tracing out locations along the trajectory. This leads to a Recurrent ResNet structure \cite{salman2018deep,cheng2021one} over the velocity MLP, where the residuals (\ie velocities) are weighted by factor $\Delta t$ before added back to the previous $uv$ coordinates\footnote{In case the time interval is not a multiple of $\Delta t$, one may simply reduce the last step size accordingly.}, as illustrated in Figure~\ref{fig:mnet}. On a side note, despite the discretization, Equation~(\ref{eq:discrete}) is still provably diffeomorphic as long as $\Delta t$ is no greater than the inverse Lipschitz constant of velocity MLP $V$.

\section{Loss definition and implementation details}

\subsection{Training loss}
We train our neural video representations with an RGB loss as well as regularization terms.

\paragraph{RGB loss}
The RGB loss is defined as the $\ell^1$ distance between the original and reproduced videos.
\begin{equation}
    L^{RGB} = \sum_{u,v,t} |I(u,v,t) - \hat{I}(u,v,t)|_1
\end{equation}
Here $I$ and $\hat{I}$ denote the reproduced and input RGB values. We normalize the input video pixels' $u,v,t$ to $[-1,1]$, and set the canonical frame's timestamp to $t_F=0$. Note that the canonical frame may still expand outside the region of $[-1,1]^2$ to preserve contents that move out of the video. 

\paragraph{Velocity regularization}
We encourage video dynamics to be explainable by as small and smooth motions as possible. This translates to a penalty on velocities
\begin{equation}
    L^{V} = \sum_{u,v,t,i} \|(\mathcal{I}-\alpha \nabla^2_{u,v})V_i(u,v,t)\|^2.
\end{equation}
Here $\mathcal{I}$ and $\nabla^2_{u,v}$ denotes the identity mapping and the Laplacian operator with respect to $u,v$, and $alpha\ge0$ is a weight that controls the spatial smoothness of velocities.

An additional `inertia loss' is applied so that video motions tend to preserve their velocity along any trajectory. Mathematically this loss is defined as the variance of velocities along any trajectory $C(u,v)$ that starts from $(u,v,t=0)$
\begin{align}
    L^{I} &= \sum_{u,v,i} \mathbf{Var}\Big( V_i\big(M(u,v,0,t),t\big)\Big),\\&\text{where}\;t\sim \mathbf{Uniform}\nonumber.
\end{align}
We compute this loss by uniformly sample $t$ during the video length.

\paragraph{Overall loss}
The overall loss function is a weighted sum of individual losses.
\begin{equation}
    L = L^{RGB} + \lambda_V L^{V} + \lambda_I L^{I} 
\end{equation}

The total complexity for computing loss functions scales linearly with video resolution and number of frames. 

\subsection{Parameters and training details}
In all experiments, the image coordinates and timestamps are all normalized to range $[-1,1]$, and we select $t_F=0$ to be the canonical frames timestamp. Networks are trained jointly with Adam optimizer for 400 epochs, and with batch size of 4,096 video pixels per batch. Training typically takes a few hours depending on input video size.

We weight the regularization losses according to number of frames available in the input video clip. When there are multiple (3 or more) frames, the video offers sufficient constraint on the motion trajectories, and we use a small weight for regularization terms with $\lambda^V=\lambda^I=0.01$ and $\alpha=0$ and set the step size to $\Delta t=0.02$. Conversely, when there are only 2 frames available, we set $\Delta t=0.2$ and increase the regularization $\lambda^V=\lambda^I=10$ and $\alpha=0.5$ to encourage a more linearized motion fields.

\section{Experiments}
\subsection{Overview}
In this section we provide detailed experiment results and comparisons with state-of-the-art methods. Since our method receives only a single video clip as input, it is data-agnostic. For this reason it is validated without a large scale test set. We provide evaluations on the Middlebury dataset \cite{baker2011database} and select 5 videos each from the Vimeo90K \cite{xue2019video}, UCF101 \cite{soomro2012ucf101,liu2017video} and Adobe240fps \cite{su2017deep} datasets.

\subsection{Comparisons}

\begin{table*}[!htb]
    \caption{\textbf{Comparison of quantitative results by different methods.} Unsupervised methods are marked with *. Our solution outperforms the other unsupervised method by Reda \etal \cite{reda2019unsupervised} and is comparable to fully supervised methods that require a large database for training. Shown are the absolute interpolation error (AIE), peak signal to noise ratio (PSNR) and structural similarity (SSIM) between predicted and ground truth images. }
    \label{tab:middlebury}
    \centering
    \subcaption*{Tested on Middlebury-other dataset with 13 video clips \cite{baker2011database}}
    \begin{tabular}{c|*{4}{|c}|*2{|c}}
        Metrics & DAIN\cite{bao2019depth} & TOFlow\cite{xue2019video} & ABME\cite{park2021asymmetric} & RIFE\cite{huang2020rife} & CC*\cite{reda2019unsupervised} & Ours*\\\hline
         AIE $\downarrow$  &  4.53 & 3.36 & 2.36 & 2.76 & 3.92 & 2.92\\
         PSNR $\uparrow$ & 29.6 & 32.1 & 36.6 & 34.2 & 30.4 & 33.8 \\
         SSIM $\uparrow$ & 0.94 & 0.96 & 0.98 & 0.97 & 0.95 & 0.97 \\
    \end{tabular}
    \subcaption*{Tested on 5 videos selected from the UCF101 dataset \cite{soomro2012ucf101,liu2017video}. }
    \label{tab:ucf101}
    \centering
    \begin{tabular}{c|*{4}{|c}|*2{|c}}
        Metrics & DAIN\cite{bao2019depth} & TOFlow\cite{xue2019video} & ABME\cite{park2021asymmetric} & RIFE\cite{huang2020rife} & CC*\cite{reda2019unsupervised} & Ours*\\\hline
         AIE $\downarrow$ & 4.77 & 3.74 & 3.19 & 3.19 & 4.82 & 3.49  \\
         PSNR $\uparrow$ & 29.8 & 32.7 & 34.5 & 34.3 & 28.9 & 32.8  \\
         SSIM $\uparrow$ & 0.96 & 0.97 & 0.98 & 0.98 & 0.96 & 0.97  \\
         
    \end{tabular}
    
    \subcaption*{Tested on 5 videos selected from the Vimeo90k dataset \cite{xue2019video}. }
    \label{tab:ucf101}
    \centering
    \begin{tabular}{c|*{4}{|c}|*2{|c}}
        Metrics & DAIN\cite{bao2019depth} & TOFlow\cite{xue2019video} & ABME\cite{park2021asymmetric} & RIFE\cite{huang2020rife} & CC*\cite{reda2019unsupervised} & Ours*\\\hline
         AIE $\downarrow$ & 3.65 & 2.81 & 1.96 & 2.23 & 3.56 & 2.81 \\
         PSNR $\uparrow$ &  31.8 & 33.4 & 37.1 & 35.4 & 31.5 & 33.0 \\
         SSIM $\uparrow$ &  0.96 & 0.97 & 0.99 & 0.98 & 0.96 & 0.96 \\
         
    \end{tabular}
    \vspace*{-1em}
\end{table*}

Table~\ref{tab:middlebury} lists the error metrics on different datasets. Due to the training time and data-agnostic nature of our method, we compare on the complete Middlebury database \cite{baker2011database} as well as 5 videos from UCF101 dataset \cite{soomro2012ucf101,liu2017video} and Vimeo90k dataset \cite{xue2019video} each (we select the first 5 videos from UCF101 and Vimeo90k with different types of scenes). Our method outperforms the other unsupervised approach by Reda \etal \cite{reda2019unsupervised} by a large margin, and performs comparably to fully supervised state-of-the-arts.

\begin{figure}[!ht]
\centering
\includegraphics[width=\textwidth]{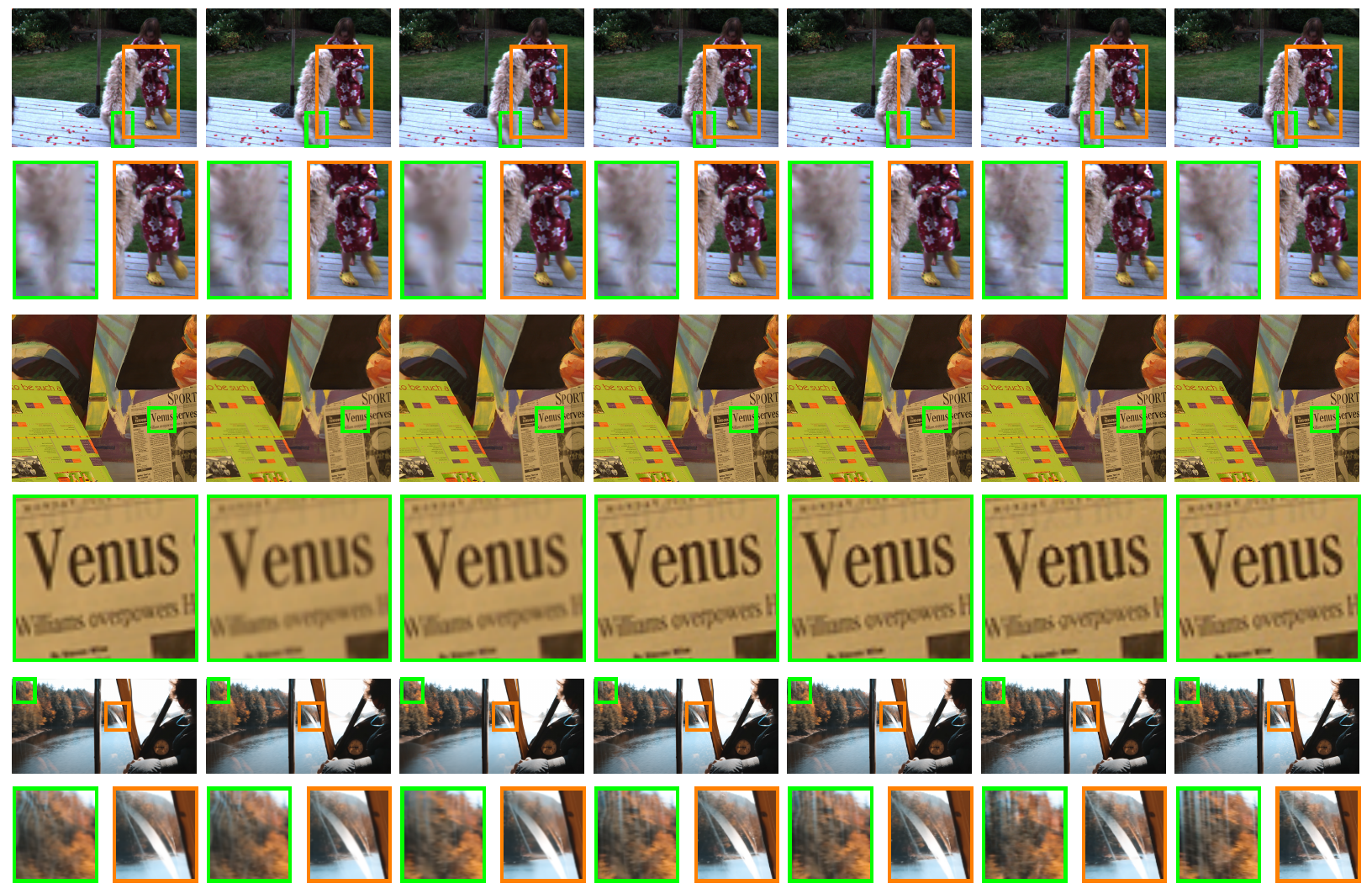}
\input{Comparisons/fig_comparison_combined}
\caption{\textbf{Comparison of qualitative results by different methods.} While ABME~\cite{park2021asymmetric} outperforms the rest methods quantitatively, it tends to blur texture details while DAIN~\cite{bao2019depth}, RIFE~\cite{huang2020rife} CC~\cite{reda2019unsupervised} and our method were able to preserve them. In the bottom row zoom-ins, our layer-based approach is able to separate reflections on the glass window from the background while other methods either distort the reflections or blur the background.}\label{fig:middlebury}
\vspace*{-1em}
\end{figure}

Figure~\ref{fig:middlebury} illustrates qualitative results of different methods on two example videos. We observe that despite ABME \cite{park2021asymmetric} consistently achieves the best quantitative performance, it tends to blur texture details in the zoom-ins. On the other hand, RIFE~\cite{huang2020rife} and our method are able to retain the original contrast and fidelity of videos.

\subsection{Ablation study and more results} \label{section:resuslts}

\begin{figure}[!ht]
\centering
\input{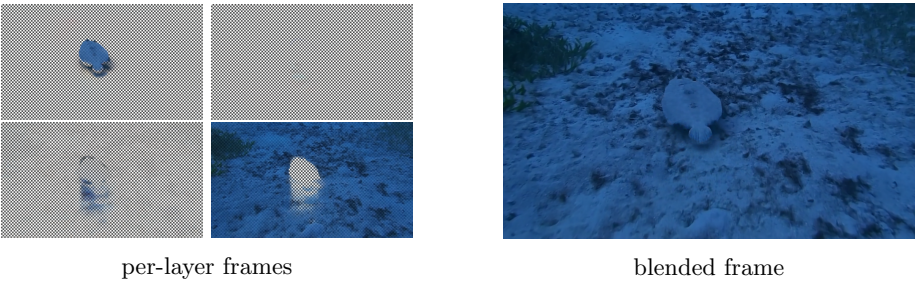}
\caption{\textbf{Layer-wise separation of a camouflaged animal.} The neural networks learn to split the flatfish and the background into different layers solely from the input frames. The top left layer contains the fish and its shadow while the background is visible in the bottom right layer.}\label{fig:fish}
\vspace*{-1em}
\end{figure}

\paragraph{Layer-wise separation on a camouflaged video}
We test our method on a camouflaged animal video from the MoCA dataset~\cite{lamdouar2020betrayed}. Our purpose is to show that the foreground animal is successfully separated from the moving background using motion cues. Our networks are trained on the video without additional inputs, and are able to tell apart the fish along with its shadow from the background despite its camouflaged appearance. Figure \ref{fig:fish} illustrates the layer-wise RGB frames weighted by corresponding visibility maps.

\begin{figure}[!htb]
    \centering
    \input{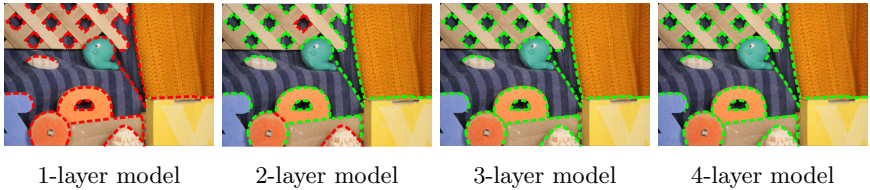}
    \caption{\textbf{Human classified occlusion boundaries with different number of video layers.} We train 1-layer, 2-layer, 3-layer and 4-layer models on the same \textit{RubberWhale} sequence and manually mark the occlusion boundaries on the interpolated frames where there is relative motion between the occluder and the occludee. The green dotted lines indicate occlusion boundaries that have been successfully set apart into different video layers by our method. Red dotted lines are boundaries that were not separated due to limited number of video layers. A single layer model is unable to handle occlusions, while a 3-layer model suffices to account for all occlusion boundaries in the video.}
    \label{fig:occlusion_boundaries}
    \vspace*{-1em}
\end{figure}

\paragraph{Separation with different number of layers}
Here we validate our layer-based occlusion handling approach with varying number of video layers. We train our method with 1-layer, 2-layer, 3-layer, 4-layer and 6-layer models, and manually verify whether the occlusion boundaries have been successfully split to different layers. Figure~\ref{fig:occlusion_boundaries} shows the results on interpolated frames, and Table~\ref{tab:no_frames} lists the corresponding errors. We note that a 3-layers model already accounts for all the motion occlusions in the example video, while in all other experiments we use a 4-layer model by default.

\begin{figure}[!htb]
    \centering
    \input{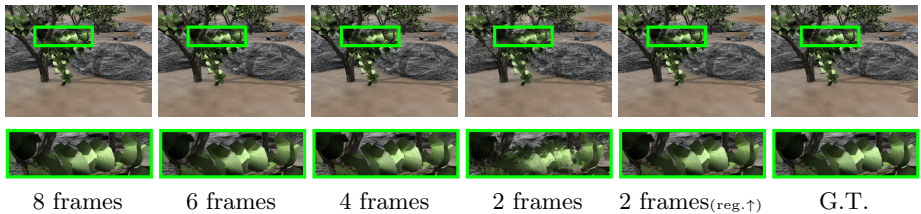}
    \caption{\textbf{Frame interpolation with different number of input frames and increased regularization.} Tested on the \textit{Grove2} sequence. When only 2 frames are available, increased regularization leads to substantially better image quality.}
    \label{fig:grove}
    \vspace*{-3em}
\end{figure}

\begin{table}
    \centering
    \caption{\textbf{Performance versus number of video layers.} The performance steadily improves with increasing number of layers when using 4 or less layers.}
    \label{tab:no_frames}
    \begin{tabular}{c|c|c|c|c|c}
     & 1 layer & 2 layers & 3 layers & 4 layers & 6 layers\\\hline
        AIE $\downarrow$ & 1.94 & 1.49 & 1.32 & 1.21 & 1.29  \\
         PSNR $\uparrow$ & 38.4 &  41.5 & 42.9 & 43.5 & 43.5 \\
         SSIM $\uparrow$ & 0.98 &0.99 & 0.99 & 0.99 & 1.00 
    \end{tabular}
    \vspace*{-1em}
\end{table}
\paragraph{Number of input frames and regularization losses}
We evaluate the frame interpolation accuracy with respect to number of input frames in the video. For this purpose we fix the regularization weights at $\lambda_V\lambda_I=0.01$ and run our method on the same sequence in Middlebury dataset \cite{baker2011database} with a total of 8, 6, 4 and 2 frames, removing 2 input frames from both ends at a time. Table \ref{tab:no_frames} shows how performance changes with decreasing number of inputs. It is observed that the performance does not degrade until only 2 input frames are provided. This is because when only 2 frames are available, the motion trajectories are governed solely by two endpoints without intermediate supervision. Lacking any better prior knowledge for video motions, we increase the regularization weights to $\lambda_V\lambda_I=10$ and step size to $\alpha=0.5$. This disambiguates motion fields by encouraging video motions to be as linear as possible, and significantly improves interpolation quality (last column of Table \ref{tab:no_frames}). Figure~\ref{fig:grove} shows the interpolated frames with different number of input frames and regularization weights.

\begin{table}
    \centering
    \caption{\textbf{Performance versus number of input frames.} By increasing the regularization weights we linearize video motions. This is particularly helpful when only 2 frames are available.}
    \label{tab:no_frames}
    \begin{tabular}{c|c|c|c|c|c}
     & 8 frames & 6 frames & 4 frames & 2 frames & 2 frames (reg.$\uparrow$)\\\hline
        AIE $\downarrow$ & 2.41 & 2.45 & 2.40 & 20.4 & 3.47  \\
         PSNR $\uparrow$ & 35.7 &  35.3 & 36.1 & 17.9 & 32.7 \\
         SSIM $\uparrow$ & 0.99 &0.99 & 0.99 & 0.49 & 0.97 
    \end{tabular}
    \vspace*{-1em}
\end{table}

\section{Conclusion, limitations and future work}
In this paper we have put forward a novel unsupervised approach for video frame interpolation. We train implicit neural representation for a brightness field and a motion field. Multiple video layers are stacked together for occlusion handling, with a fully differentiable SoftMin aggregation. Thanks to the recurrent weighted residual motion network, our method is time-consistent by construction. Unlike previous learning-based approach, we only require a single input video. We also propose regularization losses to constrain the motion fields, which proved to be especially helpful when only 2 input frames are available.

Our method has a few drawbacks. Admittedly, our method is trained on a per-video basis and it took hours to process a video. On the other hand, existing methods (\eg \cite{bao2019depth,park2021asymmetric,reda2019unsupervised}), while trained a large database, only take seconds to run. Compared to recent supervised methods, our approach slightly trails behind in quantitative performance. One possible solution for speeding up our approach is via meta learning or pre-parameterization, which is known to significantly reduce training time for implicit neural networks by several orders \cite{tancik2021learned,muller2022instant}. To improve performance, one may also use an off-the-shelf optical flow pipeline (\eg GMA~\cite{jiang2021learning}) to supervise the motion fields. This could further help with large displacement video motions. Our source codes, model and data will be made publicly available to facilitate reproducible research.

{\small
\bibliographystyle{ieeetr}
\bibliography{egbib}
}

\clearpage
\newpage
\newpage

\end{document}